\documentclass{article}
\usepackage{spconf,amsmath,graphicx,hyperref}
\usepackage{tipa}
\usepackage{cite}
\usepackage{algorithm}
\usepackage{algpseudocode}
\usepackage{booktabs}
\usepackage{multirow}
\usepackage{makecell}
\usepackage{subcaption}

\title{PAC: Pronunciation-Aware Contextualized Large Language Model-based Automatic Speech Recognition}
\name{Li Fu$^*$, Yu Xin$^*$, Sunlu Zeng, Lu Fan, Youzheng Wu, Xiaodong He}
\address{JD AI Research}
\begin{document}     
\ninept
\maketitle
\def\thefootnote{*}
\footnotetext{Contributed equally to this work.}
\def\thefootnote{\arabic{footnote}}

\begin{abstract}
This paper presents a \underline{\textbf{P}}ronunciation-\underline{\textbf{A}}ware \underline{\textbf{C}}ontextualized (\textbf{PAC}) framework to address two key challenges in Large Language Model (LLM)-based Automatic Speech Recognition (ASR) systems: effective pronunciation modeling and robust homophone discrimination. Both are essential for raw or long-tail word recognition. The proposed approach adopts a two-stage learning paradigm. First, we introduce a \textit{pronunciation-guided context learning} method. It employs an interleaved grapheme-phoneme context modeling strategy that incorporates grapheme-only distractors, encouraging the model to leverage phonemic cues for accurate recognition. Then, we propose a \textit{pronunciation-discriminative reinforcement learning} method with perturbed label sampling to further enhance the model’s ability to distinguish contextualized homophones. Experimental results on the public English Librispeech and Mandarin AISHELL-1 datasets indicate that PAC: (1) reduces relative Word Error Rate (WER) by 30.2\% and 53.8\% compared to pre-trained LLM-based ASR models, and (2) achieves 31.8\% and 60.5\% relative reductions in biased WER for long-tail words compared to strong baselines, respectively.
\end{abstract}


\vspace{-0.05in}
\section{Introduction}
\vspace{-0.05in}
\label{sec:introduction}
Recently, Automatic Speech Recognition (ASR) models based on Large Language Models (LLMs) have achieved remarkable progress by leveraging extensive linguistic knowledge~\cite{fathullah2024prompting,chen2024salm,ma2025speech,chu2024qwen2,xu2025leveraging,wang2025contextasr,chen2024streaming,jia2025efficient,chen2024bestow,meng2025large,deng2025transducer,zhou2025cjst,mu2025efficient,seide2024speech,xu2025fireredasr}. Despite these advances, recognizing rare and long-tail words, such as proper nouns, personal names, and address names, remains a persistent challenge, especially when such words are infrequent or absent in the training speech corpus. Correctly transcribing these words is crucial for understanding spoken utterances~\cite{pundak2018deep,ayache2024whisperner}. To address this problem, existing approaches often incorporate contextual information, such as keyword lists or topic cues, to help ASR systems better identify these challenging terms~\cite{lakomkin2024end,gong2024contextual,bai2024seed,yang2024ctc,gong2025br,xiao2025contextual,fang2025joint,tang2024contextualized}. However, two critical limitations still hinder further improvements:

\begin{figure}[t]
    \centering
    \includegraphics[width=0.955\linewidth]{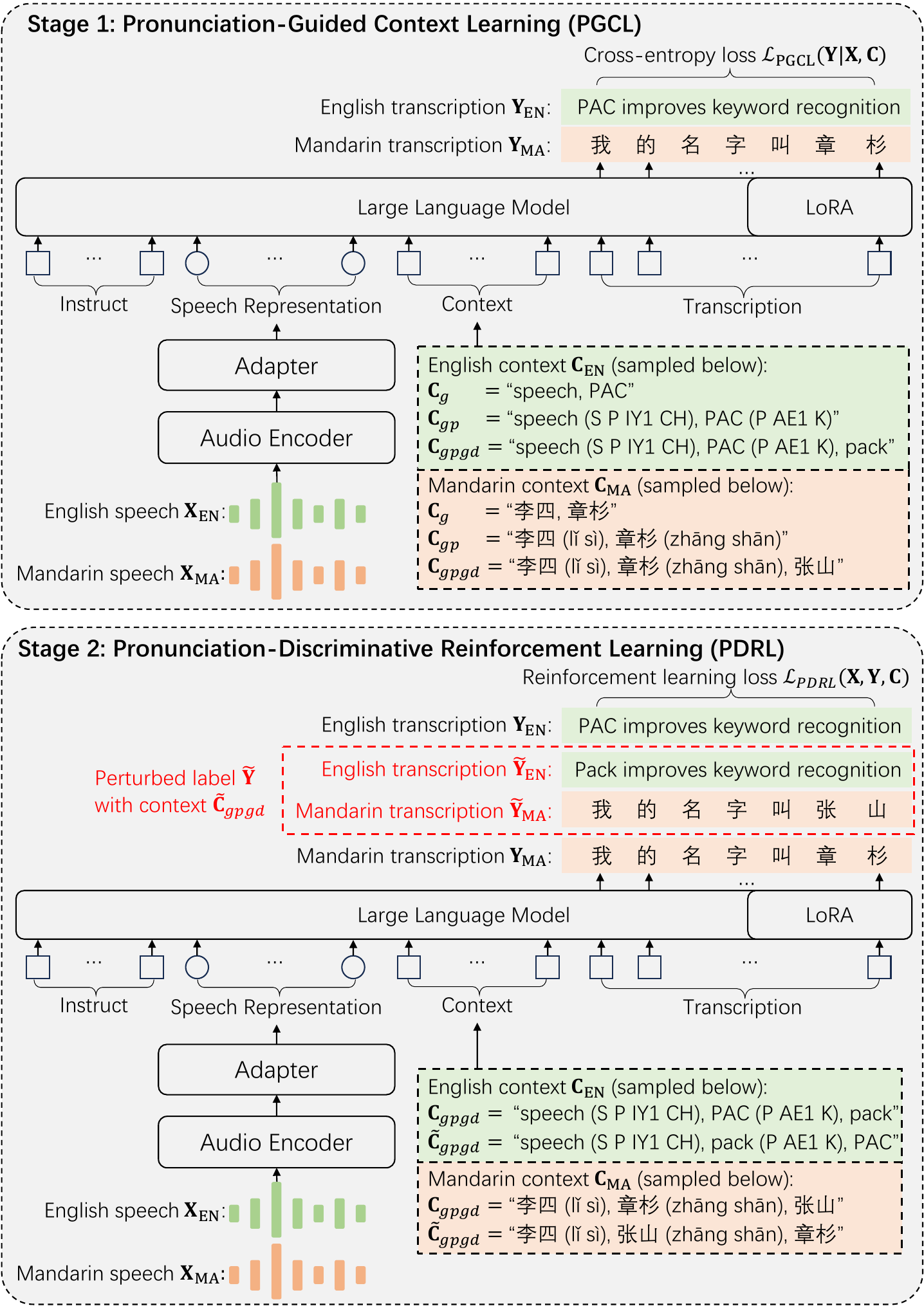}
    \caption{Overview of the proposed PAC framework for LLM-based ASR. PAC utilizes a two-stage learning paradigm, including PGCL and PDRL, to improve recognition of raw and long-tail words. Example cases in both English and Mandarin are illustrated.}
    \label{fig:fig1}
    \vspace{-0.2in}
\end{figure}

\textit{1) Lack of explicit pronunciation modeling.} Most existing contextual methods rely primarily on graphemic cues, while pronunciation information is only implicitly captured. This approach is often insufficient for handling out-of-vocabulary words, acronyms, or rare proper nouns~\cite{lei2025contextualization}. For example, consider the raw word ``psalm'' in the utterance ``Please tell me something about psalm''. This word is pronounced as \textipa{/sA:m/}, despite its spelling suggesting an initial ``p'' sound. In the absence of explicit phonemic modeling, ASR models may face difficulties in correctly recognizing such words.

\textit{2) Inadequate homophone discrimination.} Conventional training based on speech-text alignment does not explicitly guide models to distinguish between homophones or near-homophones. Even when contextual hints include the keyword (e.g., ``PAC''), the model may still favor more frequent homophones (e.g., ``pack''), particularly in Mandarin, where homophones are common~\cite{tang2024pinyin}. This bias towards familiar words reduces recognition accuracy for rare words.

To address these challenges, we propose the Pronunciation-Aware Contextualized (PAC) framework, which explicitly integrates graphemic and phonemic information into contextual modeling for LLM-based ASR. PAC employs a two-stage learning paradigm: (1) \textit{Pronunciation-Guided Context Learning (PGCL)}, which interleaves grapheme and phoneme annotations to enhance the model’s sensitivity to pronunciation cues; and (2) \textit{Pronunciation-Discriminative Reinforcement Learning (PDRL)}, which applies perturbed label sampling to strengthen the model’s ability to distinguish contextualized homophones. This dual-stage approach enables PAC to robustly recognize raw words and mitigates bias towards frequent alternatives. The main contributions of this work are listed as follows:

1) \textbf{Joint graphemic-phonemic contextual modeling:} To the best of our knowledge, this is the first work to jointly model graphemic and phonemic context in LLM-based ASR, substantially enhancing recognition of long-tail words.

2) \textbf{Pronunciation-aware training strategy:} We introduce a two-stage learning method with PGCL and PDRL, enabling explicit pronunciation modeling and robust homophone discrimination.

3) \textbf{State-of-the-art performance:} Extensive experiments on English Librispeech~\cite{panayotov2015librispeech} and Mandarin AISHELL-1~\cite{bu2017aishell} demonstrate that PAC achieves significant improvements in Word Error Rate (WER) and biased WER (B-WER) over strong baselines.

\begin{figure}[t]
\vspace{-0.4in}
\centering
\scalebox{0.8}{
\begin{minipage}{\linewidth}
\begin{algorithm}[H]
\caption{Pronunciation-Guided Context Construction for PGCL} 
\label{alg:context-construction}
\begin{algorithmic}[1]
\Require Grapheme-only context $\mathbf{C}_g$, reference transcription $\mathbf{Y}$, probabilities $P_1$, $P_2$, grapheme-to-phoneme mapping $\mathcal{T}$
\Ensure Sampled context: grapheme-phoneme $\mathbf{C}_{gp}$, grapheme-phoneme with graphemic distractor $\mathbf{C}_{gpgd}$, or  $\mathbf{C}_g$

\State Sample $r \sim \mathcal{U}(0,1)$
\If{$r < P_1$}
    \For{each word $w$ in $\mathbf{C}_g$}
        \State Add $(w, \mathcal{T}(w))$ to $\mathbf{C}_{gp}$
    \EndFor
    \State \Return $\mathbf{C}_{gp}$
\ElsIf{$r < P_1 + P_2$}
    \For{each word $w$ in $\mathbf{C}_g$}
        \If{$w$ in $\mathbf{Y}$}
            \State Generate homophone distractor $w'$ for $w$
            \State Add $(w, \mathcal{T}(w), w')$ to $\mathbf{C}_{gpgd}$
        \Else
            \State Add $(w, \mathcal{T}(w))$ to $\mathbf{C}_{gpgd}$
        \EndIf
    \EndFor
    \State \Return $\mathbf{C}_{gpgd}$
\Else
    \State \Return $\mathbf{C}_g$
\EndIf
\end{algorithmic}
\end{algorithm}
\end{minipage}
}
\vspace{-0.2in}
\end{figure}

\vspace{-0.05in}
\section{Related work}
\vspace{-0.05in}
\label{sec:related_work}
\noindent \textbf{LLM-based contextual ASR.}
Recent work on LLM-based ASR has focused on using contextual information to improve recognition of rare words. Lakomkin et al.~\cite{lakomkin2024end} showed that adding auxiliary context, such as video titles and descriptions, boosts rare word accuracy in LLaMA-based ASR. Gong et al.~\cite{gong2024contextual} compared tagged prompts and natural language biasing, highlighting the impact of prompt design on keyword recognition. Bai et al.~\cite{bai2024seed} combined context with Minimum MWER (MWER)-based reinforcement learning for further gains. To handle large keyword lists, Yang et al.~\cite{yang2024ctc} proposed coarse Connectionist Temporal Classification (CTC) decoding for hotword filtering, and Gong et al.~\cite{gong2025br} introduced a retrieval-based framework to reduce the effect of excessive hotwords. However, these approaches mainly rely on graphemic context and lack explicit pronunciation modeling. Our method addresses this gap by interleaving grapheme and phoneme annotations, allowing the model to better utilize pronunciation cues for improved keyword recognition.

\begin{figure}[t]
    \vspace{-0.3in}
    \centering
    \includegraphics[width=0.9\linewidth]{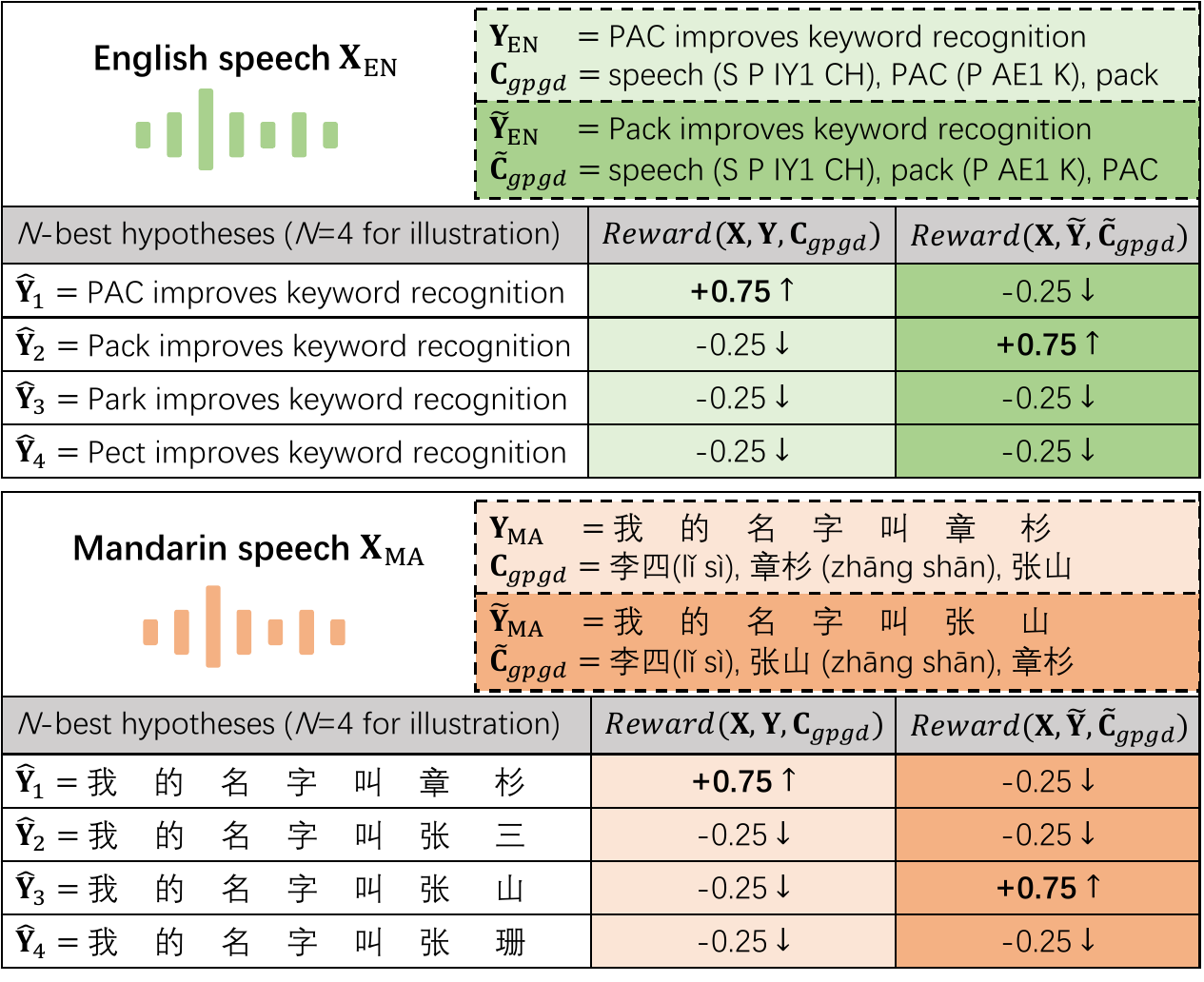}
    \caption{Overview of PDRL with perturbed transcript $\tilde{\mathbf{Y}}$ and context $\tilde{\mathbf{C}}_{gpgd}$, along with B-WER reward ($W_b(\hat{\mathbf{Y}}_i, \mathbf{Y}) - \overline{W_b}$) for homophone discrimination. Note that the N-best hypotheses sampled from the original input $(\mathbf{X}, \mathbf{Y}, \mathbf{C})$ and the perturbed input $(\mathbf{X}, \tilde{\mathbf{Y}}, \tilde{\mathbf{C}}_{gpgd})$ may diff; for illustration, they are shown in a simplified manner.}
    \label{fig:fig2}
    \vspace{-0.2in}
\end{figure}

\noindent \textbf{Phoneme-aided contextual ASR.} Phoneme embeddings have proven effective in conventional end-to-end ASR models~\cite{bruguier2019phoebe,chen2019joint,yang2024contextual}. Shi et al.~\cite{shi2024seaco} also used implicit keyword detection to filter non-top-$k$ hotwords, yielding notable improvements in Mandarin ASR. However, phoneme-aided approaches for LLM-based ASR remain underexplored. For instance, Lei et al.~\cite{lei2025contextualization} introduced a phonetic retrieval-based augmentation for LLM-based ASR, but its multi-stage pipeline introduces extra latency and complexity. In contrast, our method adopts a simple yet effective two-stage training paradigm that directly incorporates phonemic cues into LLMs, enabling efficient and accurate keyword recognition.

\section{OUR PROPOSED APPROACH}
\label{sec:our_method}
\subsection{Overview of PAC}
To improve recognition of rare words, a pre-trained LLM-based ASR model $\mathcal{M}_p$ is adapted into a contextualized model $\mathcal{M}_c$. As illustrated in Fig.~\ref{fig:fig1}, this process leverages training data triples $\{\mathbf{X}, \mathbf{C}, \mathbf{Y}\}$, where $\mathbf{X}$ denotes the input speech signal, $\mathbf{C}$ provides contextual information, and $\mathbf{Y}$ is the reference transcription. The model architecture consists of an LLM backbone guided by instruction prompts and context, along with an adaptor module that maps encoded speech representations to the textual modality.

The core of PAC is a pronunciation-aware training paradigm that explicitly incorporates both graphemic and phonemic context information. This increases the model's sensitivity to pronunciation variations and improves keyword recognition. The paradigm has two stages. First, PGCL uses supervised fine-tuning to construct interleaved grapheme-phoneme contexts and introduce distractor words, encouraging the model to leverage phonemic cues. Second, PDRL applies reinforcement learning with perturbed label sampling, exposing the model to challenging homophone scenarios and strengthening its discrimination capability. Details are provided as follows.

\begin{table}[t]
\vspace{-0.2in}
\caption{Comparison of PAC and baseline methods on English LibriSpeech test sets. ``NA'' indicates no biasing keywords, ``GT'' refers to ground-truth keyword biasing, and $N$ is the bias list size.}
\label{tab:librispeech_results}
\resizebox{0.48\textwidth}{!}{
\begin{tabular}{lcccccc}
\toprule
\multirow{2}{*}{Setting} 
    & \multicolumn{2}{c}{BR-ASR~\cite{gong2025br}} 
    & \multicolumn{2}{c}{CFL~\cite{yang2024ctc}} 
    & \multicolumn{2}{c}{PAC (Ours)} \\
\cmidrule(lr){2-3} \cmidrule(lr){4-5} \cmidrule(lr){6-7}
& WER & B-WER & WER & B-WER & WER & B-WER \\
\midrule
\multicolumn{7}{l}{\textit{test-clean}} \\
NA         & --   & --   & 1.82 & 8.28 & \textbf{1.82} & \textbf{8.26} \\
GT         & 1.1  & 1.2  & 0.77 & 0.47 & \textbf{0.76} & \textbf{0.33} \\
$N=100$    & 1.2  & 2.3  & 0.92 & 1.72 & \textbf{0.90} & \textbf{1.30} \\
$N=500$    & 1.2  & 2.4  & 1.11 & 2.05 & \textbf{0.93} & \textbf{1.35} \\
$N=1000$   & 1.2  & 2.6  & 1.12 & 2.22 & \textbf{0.95} & \textbf{1.74} \\
$N=2000$   & 1.2  & 2.8  & 1.19 & 2.50 & \textbf{1.18} & \textbf{1.91} \\
\midrule
\multicolumn{7}{l}{\textit{test-other}} \\
NA         & --   & --   & 4.05 & 18.32 & \textbf{4.02} & \textbf{18.17} \\
GT         & 2.6  & 3.7  & 2.10 & 1.50 & \textbf{2.01} & \textbf{1.29} \\
$N=100$    & 2.7  & 6.1  & 2.45 & 4.52 & \textbf{2.35} & \textbf{3.89} \\
$N=500$    & 2.7  & 6.3  & 2.59 & 6.00 & \textbf{2.53} & \textbf{4.54} \\
$N=1000$   & 2.7  & 6.6  & 2.75 & 6.21 & \textbf{2.63} & \textbf{5.07} \\
$N=2000$   & 2.8  & 7.1  & 2.93 & 6.75 & \textbf{2.70} & \textbf{6.19} \\
\bottomrule
\end{tabular}
}
\end{table}

\subsection{Pronunciation-guided context learning}
Instead of relying solely on conventional grapheme-only context $\mathbf{C}_g$ for LLM-based ASR, we propose a pronunciation-guided context construction strategy to explicitly incorporate pronunciation information and improve keyword recognition. Specifically, each word $w$ in $\mathbf{C}_g$ is paired with its corresponding grapheme-to-phoneme transcription $\mathcal{T}(w)$, forming the interleaved grapheme-phoneme context $\mathbf{C}_{gp}$. As illustrated in Fig.~\ref{fig:fig1}, given $\mathbf{C}_g = \text{``speech, PAC''}$, we construct $\mathbf{C}_{gp} = \text{``speech (S P IY1 CH), PAC (P AE1 K)''}$, where phoneme annotations are generated using the grapheme-to-phoneme toolkit\footnote{\url{https://pypi.org/project/g2p-en/}} for English and the pypinyin toolkit\footnote{\url{https://pypi.org/project/pypinyin/}} for Mandarin.

However, our experiments show that simply providing $\mathbf{C}_{gp}$ offers limited improvement over grapheme-only context, as the model may still focus on graphemic information and fail to fully exploit pronunciation cues. To address this limitation, we introduce an augmented guided context $\mathbf{C}_{gpgd}$ by adding grapheme-only distractor words $w'$ with similar or identical pronunciations to the target keywords in the label $\mathbf{Y}$. For example, for ``speech (S P IY1 CH), PAC (P AE1 K)'', we add the homophone distractor ``pack'' to form ``speech (S P IY1 CH), PAC (P AE1 K), pack''. This structure encourages the model to actively leverage contextual phonemic information, thereby improving recognition of the target keyword ``PAC'' in $\mathbf{Y}$. Following~\cite{huang2024improving}, we employ hand-crafted linguistic rules to identify spelling alternatives with similar phonetic patterns for English, while for Mandarin, homophone replacement is performed using the pypinyin toolkit to substitute words with identical pronunciations. The construction process for pronunciation-guided context is summarized in Algorithm~\ref{alg:context-construction}, where $P_1 = P_2 = 1/3$ are set to balance the distribution of different context types during training.

To optimize the model with diverse contextual inputs, we define the Cross-Entropy (CE) losses for the aggregate context types as:
\begin{equation}
    \scalebox{0.8}{$
    \mathcal{L}_{\text{PGCL}}(\mathbf{Y}|\mathbf{X}, \mathbf{C}) =
    \underbrace{\mathcal{L}_{\text{CE}}(\mathbf{Y}|\mathbf{X}, \mathbf{C}_g)}_{\mathcal{L}_g} +
    \underbrace{\mathcal{L}_{\text{CE}}(\mathbf{Y}|\mathbf{X}, \mathbf{C}_{gp})}_{\mathcal{L}_{gp}} +
    \underbrace{\mathcal{L}_{\text{CE}}(\mathbf{Y}|\mathbf{X}, \mathbf{C}_{gpgd})}_{\mathcal{L}_{gpgd}}
    $}
    \label{eq:eq1}
\end{equation}
where $\mathcal{L}_{\text{CE}}$ denotes the standard cross-entropy loss.

\begin{table}[t]
\vspace{-0.2in}
\caption{Comparison of PAC and baseline methods on Mandarin AISHELL-1 test sets. ``NA'' indicates no biasing keywords, ``GT'' refers to ground-truth keyword biasing, and $N$ is the bias list size.}
\label{tab:aishell1_results}
\resizebox{0.48\textwidth}{!}{
\begin{tabular}{lcccccc}
\toprule
\multirow{2}{*}{Setting} 
    & \multicolumn{2}{c}{SeACo-Paraformer~\cite{shi2024seaco}} 
    & \multicolumn{2}{c}{CFL~\cite{yang2024ctc}} 
    & \multicolumn{2}{c}{PAC (Ours)} \\
\cmidrule(lr){2-3} \cmidrule(lr){4-5} \cmidrule(lr){6-7}
& CER & B-WER & CER & B-WER & CER & B-WER \\
\midrule
\multicolumn{7}{l}{\textit{test-small}} \\
NA         & 9.94 & 84.29 & 4.57 & 34.07 & \textbf{4.44} & \textbf{34.00} \\
GT         & 2.86 & 10.71 & 1.25 & 2.14 & \textbf{0.90} & \textbf{1.07} \\
$N=10$     & 2.91 & 11.07 & 1.36 & 3.93 & \textbf{1.02} & \textbf{1.79} \\
$N=100$    & 2.99 & 11.79 & 1.73 & 7.86 & \textbf{1.23} & \textbf{3.57} \\
$N=187$ & 3.28 & 13.57 & 1.86 & 8.21 & \textbf{1.44} & \textbf{5.36} \\
\midrule
\multicolumn{7}{l}{\textit{test-middle}} \\
NA         & 5.49 & 43.34 & 2.82 & 20.59 & \textbf{2.72} & \textbf{20.40} \\
GT         & 1.92 & 4.44  & 1.09 & 2.48 & \textbf{0.77} & \textbf{0.91} \\
$N=10$     & 2.00 & 5.35  & 1.10 & 2.50 & \textbf{0.88} & \textbf{1.02} \\
$N=100$    & 2.11 & 6.48  & 1.22 & 3.87 & \textbf{0.90} & \textbf{1.71} \\
$N=400$ & 2.20 & 6.71  & 1.51 & 6.03 & \textbf{1.13} & \textbf{3.07} \\
\midrule
\multicolumn{7}{l}{\textit{test-large}} \\
NA         & 5.47 & 46.26 & 2.41 & 16.94 & \textbf{2.34} & \textbf{16.72} \\
GT         & 1.54 & 3.20  & 0.91 & 1.71 & \textbf{0.88} & \textbf{0.93} \\
$N=10$     & 1.55 & 3.20  & 1.00 & 1.78 & \textbf{0.89} & \textbf{0.93} \\
$N=100$    & 1.65 & 4.27  & 1.21 & 3.99 & \textbf{1.03} & \textbf{1.78} \\
$N=600$ & 1.89 & 6.19  & 1.48 & 6.55 & \textbf{1.10} & \textbf{2.85} \\
\bottomrule
\end{tabular}
}
\vspace{-0.2in}
\end{table}

\begin{table*}[t]
\vspace{-0.2in}
\centering
\caption{Ablation study of PAC components on LibriSpeech and AISHELL-1 test sets.}
\vspace{-0.1in}
\label{tab:firered_llm_ablation}
\resizebox{.94\textwidth}{!}{
\begin{tabular}{
    l
    |cc|cc
    |cc|cc|cc
}
\toprule
\multirow{3}{*}{Method} 
    & \multicolumn{2}{c|}{LibriSpeech test-clean} 
    & \multicolumn{2}{c|}{LibriSpeech test-other} 
    & \multicolumn{2}{c|}{AISHELL-1 test-small}
    & \multicolumn{2}{c|}{AISHELL-1 test-middle}
    & \multicolumn{2}{c}{AISHELL-1 test-large} \\
 
    & \multicolumn{2}{c|}{$N=2000$} 
    & \multicolumn{2}{c|}{$N=2000$}
    & \multicolumn{2}{c|}{$N=187$}
    & \multicolumn{2}{c|}{$N=400$}
    & \multicolumn{2}{c}{$N=600$} \\
\cmidrule(lr){2-3} \cmidrule(lr){4-5} \cmidrule(lr){6-7} \cmidrule(lr){8-9} \cmidrule(lr){10-11}
& WER & B-WER & WER & B-WER 
  & CER & B-WER & CER & B-WER & CER & B-WER \\
\midrule
Pre-trained FireRed-LLM~\cite{xu2025fireredasr} & 1.69 & 8.00 & 3.69 & 17.31 & 3.12 & 24.29 & 1.66 & 12.74 & 1.60 & 12.46 \\
\midrule
+ PGCL ($\mathcal{L}_g$) & 1.19 & 2.50 & 2.93 & 6.75 & 1.86 & 8.21  & 1.51 & 6.03  & 1.48 & 6.55 \\
+ PGCL ($\mathcal{L}_g$ + $\mathcal{L}_{gp}$)  & 1.19 & 2.35 & 2.92 & 6.63  & 1.68 & 6.92  & 1.42 & 5.33  & 1.36 & 5.42 \\
+ PGCL ($\mathcal{L}_g$ + $\mathcal{L}_{gp}$ + $\mathcal{L}_{gpgd}$)  & 1.18 & 1.97 & 2.91 & 6.37 & 1.54 & 5.81  & 1.22 & 3.53  & 1.23 & 3.35 \\
+ PGCL ($\mathcal{L}_g$ + $\mathcal{L}_{gp}$ + $\mathcal{L}_{gpgd}$) + PDRL & \textbf{1.18} & \textbf{1.91} & \textbf{2.70} & \textbf{6.19} & \textbf{1.44} & \textbf{5.36}  & \textbf{1.13} & \textbf{3.07}  & \textbf{1.10} & \textbf{2.85} \\
\bottomrule
\end{tabular}
}
\end{table*}

\subsection{Pronunciation-discriminative reinforcement learning}
To further enhance the model's ability to distinguish homophones, we introduce a PDRL strategy that leverages perturbed label sampling to create more challenging training scenarios. Specifically, for each training instance, we generate a perturbed transcription $\tilde{\mathbf{Y}}$ by replacing the target keyword $w$ in $\mathbf{Y}$ with its homophone distractor $w'$. In parallel, a perturbed pronunciation-guided context $\tilde{\mathbf{C}}_{gpgd}$ is constructed by substituting the entry $(w, \mathcal{T}(w), w')$ in $\mathbf{C}_{gpgd}$ with $(w', \mathcal{T}(w'), w)$. This data augmentation exposes the model to both original and perturbed context-label pairs, thereby improving its ability to effectively discriminate between homophones.

Inspired by~\cite{bai2024seed}, we adopt MWER for biased words as the reinforcement learning objective. Among all sampled hypotheses, those with a lower B-WER than the average receive positive rewards, while those with higher B-WER are penalized (see Fig.~\ref{fig:fig2}). To promote recognition across different contexts, the overall loss is:

\begin{equation}
    \scalebox{0.9}{$
    \mathcal{L}_{\text{PDRL}}(\mathbf{X}, {\mathbf{Y}}, {\mathbf{C}}) = \mathcal{L}_{\text{b}}(\mathbf{X}, \mathbf{Y}, \mathbf{C}_{gpgd}) + \mathcal{L}_{\text{b}}(\mathbf{X}, \tilde{\mathbf{Y}}, \tilde{\mathbf{C}}_{gpgd})
    $}
\end{equation}
where the biased MWER loss is formulated as~\cite{prabhavalkar2018minimum}
\begin{equation}
    \scalebox{0.85}{$
    \mathcal{L}_{\text{b}}(\mathbf{X}, \mathbf{Y}, \mathbf{C}) = \frac{1}{N} \sum_{\hat{\mathbf{Y}}_i \in \mathcal{N}(\mathbf{X}, \mathbf{C})} P(\hat{\mathbf{Y}}_i|\mathbf{X}, \mathbf{C}) \left[ W_b(\hat{\mathbf{Y}}_i, \mathbf{Y}) - \overline{W_b} \right]
    $}
\end{equation}
Here, $\mathcal{N}(\mathbf{X}, \mathbf{C})$ denotes the set of $N$-best hypotheses generated from the input $\mathbf{X}$ and context $\mathbf{C}$, $P(\hat{\mathbf{Y}}_i|\mathbf{X}, \mathbf{C})$ is the normalized likelihood of each hypothesis, $W_b(\hat{\mathbf{Y}}_i, \mathbf{Y})$ is the B-WER between the hypothesis and the reference, and $\overline{W_b}$ is the average B-WER over all sampled hypotheses. In our experiments, we set $N=8$. We also apply the aggregated CE loss $\mathcal{L}_{\text{PGCL}}$ with a weight of 0.01 to prevent training divergence~\cite{prabhavalkar2018minimum}.

\section{Experiments and Discussion}
\label{sec:exp_and_discussion}
\subsection{Experimental setup}
\label{subsec:exp_setup}
\noindent \textbf{Data preparation.} To validate the effectiveness of our proposed approach, we conduct experiments on both English and Mandarin datasets. For training, we use the 960-hour Librispeech corpus~\cite{panayotov2015librispeech} for English and the 170-hour AISHELL-1 corpus~\cite{bu2017aishell} for Mandarin to adapt pre-trained LLM-based ASR models with contextual information. For each training utterance, keywords are randomly selected from the reference label, and the context is constructed by combining the keywords with a randomly sampled number of arbitrary words, ranging from 1 to 100~\cite{pundak2018deep}. For evaluation, we adopt the Librispeech test-clean and test-other sets for English, following the artificial biasing list protocol from previous work~\cite{le2021contextualized}: the 5000 most frequent words in the Librispeech training set are considered common, while all others are regarded as rare. For Mandarin, we use the publicly available AISHELL-1 test subsets~\cite{shi2024seaco}: test-small, test-middle, and test-large, which contain 235, 808, and 1,334 utterances, respectively, with corresponding keyword lists of 187, 400, and 600 words.

\noindent \textbf{ASR models.\footnote{Our code and models will be released upon publication.}} We employ the open-source FireRed-LLM model~\cite{xu2025fireredasr} as our pre-trained LLM-based ASR backbone, which contains 7 billion parameters and demonstrates outstanding performance on both English and Mandarin speech recognition tasks. Further architectural details can be found in~\cite{xu2025fireredasr}. Note that our method can be readily applied to other LLM-based ASR models.

\noindent \textbf{Training and decoding.} All models are trained on 8 NVIDIA G2 140G GPUs, with each GPU processing 8000 seconds of speech per batch. The encoder and adapter modules are fully trainable, while fine-tuning of the LLM is performed using LoRA~\cite{hu2022lora}, with the same configuration as the pre-trained model. The Adam optimizer is used with a learning rate of $2 \times 10^{-5}$, warmed up over the first 1000 steps. During decoding, beam search with a beam size of 4 is applied. To mitigate hallucination issues caused by long contexts, we follow~\cite{yang2024ctc} and train an independent CTC module from the encoder, which is used for phonetic matching to filter out irrelevant keywords.

\noindent \textbf{Baselines.} We compare the proposed PAC method with several strong baselines, including BR-ASR~\cite{gong2025br} and SeACo-Paraformer~\cite{shi2024seaco}, which are recognized as leading contextual ASR models for English and Mandarin. In addition, we implement a contextual adaptation~\cite{yang2024ctc} based on FireRed-LLM (CFL) for comparison.

\noindent \textbf{Evaluation metrics.} We evaluate the overall effectiveness of our approach using WER for the English test sets and Character Error Rate (CER) for the Mandarin test sets. To specifically assess keyword recognition performance under contextual biasing, we report B-WER for all keyword-related experiments~\cite{le2021contextualized}.

\begin{figure}[t]
    \centering
    \includegraphics[width=1.0\linewidth]{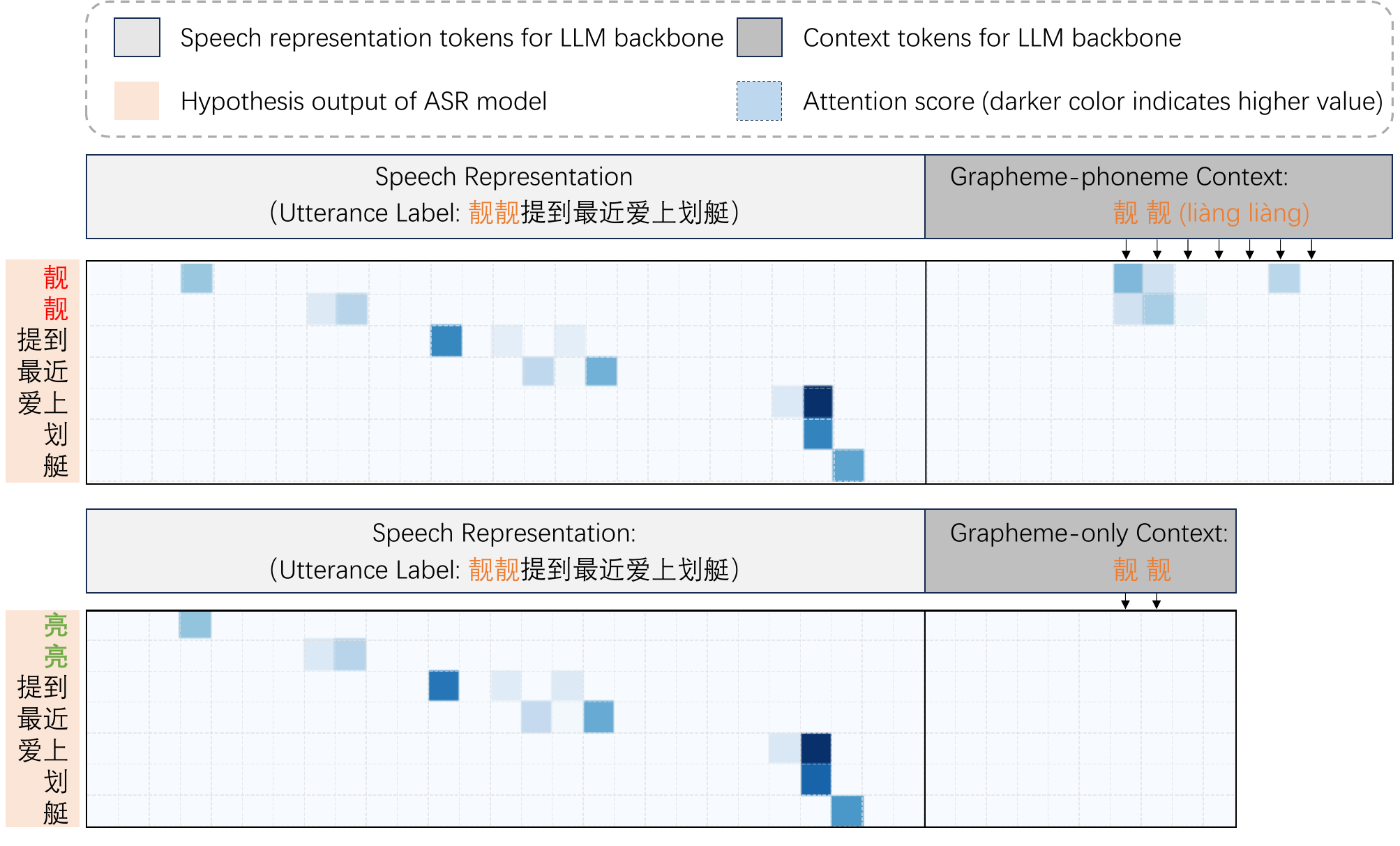}
    \caption{Attention scores from the final LLM layer. PAC leverages both graphemic and phonemic context to identify target keywords (red in hypothesis), whereas the grapheme-only context leads to recognition errors (green in hypothesis).}
    \label{fig:fig3}
    \vspace{-0.1in}
\end{figure}

\subsection{Main experiments on English and Mandarin datasets}
\label{subsec:main_exp}
Tables~\ref{tab:librispeech_results} and~\ref{tab:aishell1_results} summarize the performance of PAC compared to baselines on both Librispeech and AISHELL-1 test sets under various biasing conditions and keyword list sizes $N$. Across all scenarios, PAC consistently achieves the lowest error rates. Specifically, on the Librispeech test-clean set with $N=2000$, our approach reduces the relative B-WER by 31.8\% and 23.6\% compared to BR-ASR~\cite{gong2025br} and CFL~\cite{yang2024ctc}, respectively. On AISHELL-1, PAC delivers even greater improvements over SeACo-Paraformer~\cite{shi2024seaco} and CFL~\cite{yang2024ctc}, reducing the relative B-WER for keywords by 60.5\% and 34.7\% on the test-small ($N=197$) set. These results demonstrate that our method, by explicitly incorporating pronunciation information and leveraging the proposed two-stage training paradigm, can significantly boost keyword recognition under contextual biasing. Furthermore, we observe that the improvements are even more pronounced on Mandarin datasets. We attribute these gains to the greater occurrence of homophones in Mandarin, where pronunciation cues are more critical for distinguishing contextual words.

\subsection{Ablation study and discussion}
\label{subsec:ablation}
\noindent \textbf{Performance of PGCL.} As shown in Table~\ref{tab:firered_llm_ablation}, we evaluate the effectiveness of PGCL by training the model with various context configurations as defined in Eq.~\ref{eq:eq1}. The results demonstrate that incorporating the grapheme-phoneme with grapheme distractor loss $\mathcal{L}_{gpgd}$ yields substantial improvements compared to using only the grapheme loss $\mathcal{L}_g$ or the grapheme-phoneme loss $\mathcal{L}_{gp}$. These findings indicate that our pronunciation-guided context construction effectively encourages the model to utilize phonemic information, leading to more accurate keyword recognition.

\noindent \textbf{Performance of PDRL.} We assess the impact of PDRL by enabling it on top of PGCL (see Table~\ref{tab:firered_llm_ablation}), which results in additional error rate reductions across all datasets. Specifically, we observe a relative B-WER reduction of 3\% on LibriSpeech test-clean and 14.9\% on AISHELL-1 test-large. These results demonstrate that PDRL notably enhances the model’s ability to distinguish homophones and rare keywords. Overall, PAC achieves up to 30.2\% and 53.8\% relative reductions in WER and CER compared to the pre-trained FireRed-LLM~\cite{xu2025fireredasr} on English and Mandarin datasets, respectively.

\noindent \textbf{Analysis of pronunciation-aware modeling via attention scores.} To gain deeper insight into the mechanism of our approach, we visualize the attention scores of the LLM backbone during inference. As shown in Fig.~\ref{fig:fig3}, when provided with grapheme-phoneme context, the model assigns higher attention scores to phonemic cues associated with target keywords, especially in challenging homophone scenarios where grapheme-only models fail. This demonstrates that PAC leverages pronunciation information to guide the recognition process, thereby improving both accuracy and robustness.

\section{Conclusions}
\label{sec:conclusion}
In this paper, we present PAC, a pronunciation-aware contextualized framework for LLM-based ASR. By explicitly integrating graphemic and phonemic contextual information through a two-stage learning paradigm consisting of PGCL and PDRL, PAC effectively addresses the challenges of raw word recognition and homophone discrimination. Extensive experiments on both English and Mandarin benchmarks demonstrate that PAC consistently outperforms strong baselines, achieving substantial reductions in CER/WER and B-WER, particularly under challenging keyword biasing scenarios. In future work, we plan to extend our approach to multilingual and code-switching ASR tasks and to explore more advanced pronunciation-aware modeling strategies.

\vfill\pagebreak

\bibliographystyle{IEEEbib}
\bibliography{refs}

\end{document}